\documentclass{article}
\usepackage{INTERSPEECH2019,amsmath,graphicx,hyperref,subfig}

\title{Challenging the Boundaries of Speech Recognition: The  MALACH Corpus}

\name{Michael Picheny, Z\'{o}ltan T\"{u}ske, Brian Kingsbury, Kartik Audhkhasi, Xiaodong Cui, George Saon}
\address{IBM Research AI \\
         IBM T. J. Watson Research Center
         Yorktown Heights, NY 10598}
\email{\{picheny,Zoltan.Tuske,bedk,kaudhkha,cuix,gsaon\}@us.ibm.com}
\begin{document}
\ninept
\maketitle
\begin{abstract}
There has been huge progress in speech recognition over the last
several years. Tasks once thought extremely difficult,
such as SWITCHBOARD, now approach levels of human performance.
The MALACH corpus (LDC catalog LDC2012S05), a 375-Hour subset of a large archive
of
Holocaust
testimonies collected by the Survivors of the Shoah Visual History
Foundation, presents significant challenges to the speech
community. The collection consists of unconstrained, natural speech filled with
disfluencies, heavy accents, age-related coarticulations, un-cued
speaker and language switching, and emotional speech - all still open
problems for speech recognition systems.  Transcription is
challenging even for skilled human annotators. 
This paper proposes that the community place focus on the 
MALACH corpus to develop
speech recognition systems that are more robust with respect to
accents, disfluencies and emotional speech. 
To reduce the barrier for entry, a lexicon and training and testing
setups have been created and baseline results using current deep learning
technologies are presented. The metadata has just been released by LDC
(LDC2019S11). It is hoped that this resource will enable the 
community to build on top of these baselines so
that the extremely important information in these and related oral
histories becomes accessible to a wider audience.
\end{abstract}


\section{Introduction}
\label{sec:intro}
There has been huge progress in speech recognition over the last
several years. Tasks previously considered merely hard, such as open vocabulary
voice search and voice messaging, are now in wide deployment across
popular consumer devices such as smartphones \cite{chiu2018state} and
smart speakers \cite{raju2018contextual}. Tasks once thought extremely difficult,
such as SWITCHBOARD, have now approached levels of human performance
\cite{saon2017english,stolcke2017comparing}. The casual public now believes speech
recognition is a solved problem.
It is a fair
question to ask what problems remain unsolved in the speech
recognition area, and what research is there left to perform.

In  \cite{picheny2017TSD} it is argued that
there are many areas in which speech recognition systems still lack
robustness, especially when compared to levels of human
performance. Some of these areas include accented speech, highly
disfluent speech, and emotional speech.
A major difficulty lies in the lack of
appropriate publicly available speech recognition corpora. The
community 
evaluates on SWITCHBOARD, Wall Street Journal, and Librispeech \cite{paul1992design,panayotov2015librispeech} because the data is easy
to obtain (i.e., relatively minor or no cost to access); they have few
or no usage restrictions (e.g. effectively limited
to educational institutions or evaluation participants); and there are well documented and defined setups of
training and test data accompanied by easy-to-duplicate speech
recognition baselines.

There do exist public corpora that exhibit one or
more of these various phenomena. For example, there are disfluencies
in SWITCHBOARD, accented speech in the Mozilla corpus \cite{mozilla2017},
and very informal speech in the AMI corpus
\cite{carletta2006announcing}. However, very few corpora demonstrate
all these phenomena - SWITCHBOARD is
relatively accent free, the Mozilla Corpus is read speech, etc. The
1996/1997 LDC releases of the Broadcast News corpus \cite{woodland1997development} 
classified the data across a variety of acoustic conditions, including
non-native speech, but the total amount of such speech was quite
small. There are certainly other very diverse corpora, such as the 
corpus that comprised the MGB \cite{bell2015mgb} challenge, but the usage
license was accompanied by a number of restrictions. Of course, large industrial organizations have huge labelled
databases but these are not available in any fashion to the community.

The goal of the NSF-Sponsored MALACH project
\cite{byrne2004automatic}
was to develop
techniques to automate searching of large spoken archives. The underlying
spoken archive, the  Visual History Archive\textregistered~was created by Steven Spielberg's The
Survivors of the Shoah Visual History Foundation (VHF) \cite{vhf2018}. It was founded to preserve the stories of survivors
and witnesses of the Holocaust. It had  created what still remains to
be the largest collection of digitized oral history interviews on a
single subject: almost 55,000 interviews in 42 languages, a total
of 115,000 hours of audio and video.

\begin{figure}[t]
  \centering
  {\includegraphics[width=1.0\columnwidth]{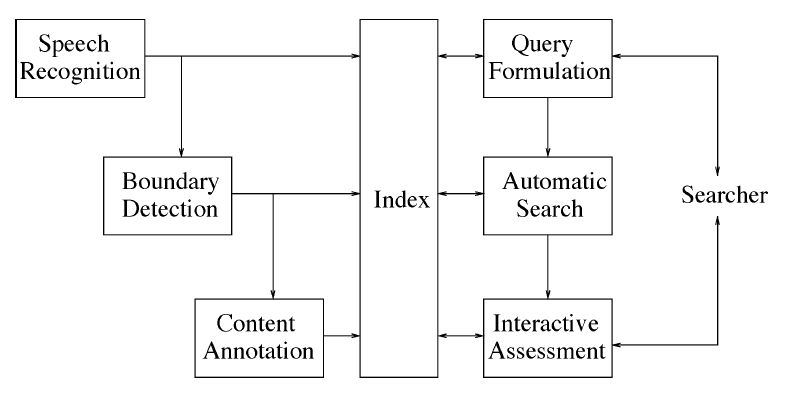}
  }
  \caption{Large Spoken Archive Search System Architecture proposed in
  MALACH \cite{byrne2004automatic}}
  \label{fig:Block}
  \vspace{-.5cm}
\end{figure}

In order to automate the creation of a large spoken searchable
archive, a high-level architecture was proposed
\cite{byrne2004automatic} and is shown in Figure~\ref{fig:Block}. It
can be seen that accurate speech recognition is the linchpin of such a
system. Without accurate speech recognition, concepts of interest cannot
be located, annotated, and searched, and all but the simplest types of
queries will fail.

Although half of the collection is in English, the testimonies were
collected from survivors whose native language was not English. The
collection consists of unconstrained, natural speech filled with
disfluencies, heavy accents, age-related coarticulations, un-cued
speaker and language switching, and emotional speech. Although the
recordings consist of relatively high-quality audio, transcription is
challenging even for skilled human annotators. The speech recognition
challenge is therefore obvious. When first proposed as a research
project, the speech
recognition task was thought to be nearly impossible. As will be
discussed, accurate recognition still remains a challenge even to today's highly
sophisticated systems. 

Several hundred hours of English testimonies were provided to the
members of the project team to build speech recognition systems and
to also serve as a basis for experiments in search technologies.
In 2012, approximately 375 hours of English testimonies
were released to the Linguistic Data Consortium (LDC) \cite{LDC2012}  along with
human transcripts for approximately 200 hours of data so that
researchers could use this data to study phenomena ranging from speech
recognition performance to socio-linguistic phenomena. (A similar
Czech corpus was released in 2014 \cite{LDC2014}). Unfortunately,
training and testing setups were not defined, limiting uptake by the
broader speech community.

This paper proposes that
the MALACH corpus be utilized to study ways of making
speech recognition systems more robust with respect to accents, disfluencies
and emotional speech.
Training and testing
setups, a lexicon, and a scoring file are described (and just released
by LDC \cite{LDC2019}). Baseline results using current deep learning
technologies are also presented. 
The hope is that this will enable the community to more easily
pick up the work to make advances in speech recognition so
that the extremely important information in these and related oral
histories becomes easily accessible.

The rest of the paper is broken up as
follows. Section~\ref{sec:prior} reviews earlier speech recognition
results on MALACH, Section~\ref{sec:release} describes basic training
and test setups. Section~\ref{sec:baseline} presents details of the baseline
systems, Section~\ref{sec:additional} describes some additional modeling
improvements, 
and Section~\ref{sec:discussion} suggests future work.

\section{Prior Speech Recognition Results}
\label{sec:prior}

The MALACH project ran from 2001-2006. As that time, the dominant speech
recognition technology was speaker-adaptive processing \cite{gales98}, sometimes combined
with more sophisticated techniques such as MMI/MPE training
\cite{povey02}  or (by the end
of the project) fMPE processing \cite{povey05}. Most of these technologies have now
been subsumed by deep learning variants. It is still useful to review
some of these early recognition results on this corpus for comparison
with what will hopefully be improved results due to more recent
speech recognition technology developments.

Measured perplexities on the task range from 72-180 depending upon how
much speech was used (65/200 hours) and what and how other language
model (LM) sources were
interpolated (SWITCHBOARD/Broadcast news) \cite{ramabhadran2003towards}.

Recognition performance numbers ranged from 43.8\% for a SAT model
trained on 65 hours of speech and an interpolated language model with
SWB and Broadcast News data \cite{ramabhadran2003towards} to 38.3\%
when 200 hours of manual transcriptions
are utilized \cite{ramabhadran2005exploiting}. Even better results (32.3\%) were reported in
\cite{ramabhadran2005exploiting} but 600 hours of unlabelled data were included
and this additional data is not available in the LDC corpus.

\section{Proposed Training and Testing Partitions}
\label{sec:release}

The MALACH corpus previously released through the LDC consists of selections from 784
interviews ranging from approximately 15 minutes to 30 minutes in
length.
There were many ways in which the 784 interviews could have been
divided into training and test data. In the interest of continuity with
prior work
\cite{ramabhadran2003towards,byrne2004automatic,ramabhadran2005exploiting}
and to allow us to leverage the data preparation that was
originally performed, we decided to use as many
of the transcribed portions of the conversations that were originally used for training and testing in 
older work and were actually released as part of \cite{LDC2012}. This
yielded 674 conversations from the original 693 interviews used for
training and 8 from the original 10 interviews used for testing. 
Informal experiments comparing
error rates on the abbreviated test data relative to the original test data  did not reveal any
appreciable change in overall WER. This leaves 102 additional conversations that can be used for a
broader test set to evaluate aspects of disfluencies, emotional speech,
etc., or rolled into the training data.

Using the above split, we defined a basic training and test
set. The basic training data set consists of 176 hours of
manually transcribed speech and
the test data consists of 3.1 hours.
The basic training data as extracted from the supplied transcripts consists of
1.3M tokens. In these 1.3M tokens, approximately 44K tokens were
marked as filled pauses and 15K tokens were marked as partial words
(disfluencies). Note that sections of the interview containing highly
emotional speech were tagged by the transcriber and marked as such in
the interview. We found in the 674 training interviews 522 explicitly tagged
emotional events. The test set did not contain any such events, but the
unused (transcribed) interviews
contain 65 such events (and might be useful for some
simple experiments in emotional tagging). The proposed test
set contains 26K
tokens. Of these tokens, 93.3\% are covered by the tokens in the
training set.   

To further allow for comparison to previously reported results on
MALACH, a minitest of 1.5 hours  was created. It is identical to the test data reported on in previous MALACH
work \cite{byrne2004automatic} except
for the two conversations (four speakers) that were not
released publicly.  All results presented in this paper are on this
minitest, as the full test set is still undergoing verification. The
training vocabulary covered 98.1\% of the tokens in the minitest.

\section{Baseline Results}
\label{sec:baseline}
As described in Section~\ref{sec:prior}, the original MALACH results were based
on an older generation of speech recognition technology. 
Since
then, due to the success of deep learning, there has been a major
revolution in speech recognition. Systems today are almost
unrecognizable from those of 10-15 years ago, and technology changes
on almost a daily basis.

To better situate results in a compact historical perspective, a
set of increasingly complex systems were built ranging from basic
context-dependent state-based hidden Markov models all the way through
to LSTM-based hybrid models.

\subsection{Acoustic Processing}
\label{ssec:AP}
The interviews in the standard distribution are provided as
two-channel {MP2} files. Although two
separate microphones were used to collect data from the interviewer
and the interviewee, placement was sometimes arbitrary, channel
failures occured sporadically, and the microphones were 
sometimes switched in the middle of an interview. In most cases, the
transcriber indicated for which channel the interviewee dominated, and
that was the channel chosen for downstream processsing. For the test
data, the best channels for both the interviewee and the interviewer
were chosen manually. 

In this study, the manual segmentations
determined by the transcribers were used for both training 
purposes (these segment boundaries are
included in the MALACH distribution). For testing, the 
segmentations went through an additional pass of manual verification.  The average segment length
in the test data was 6.2 seconds. 

The interviews were provided at a sampling rate of 44.1 KHz. The channels were separated and
downsampled to 16 KHz.  The input feature space consisted of
40-dimensional logmel or PLP features after first applying global cepstral mean
and variance normalization followed by  utterance-based cepstral mean
normalization.

\subsection{Lexical and Language Modeling}
\label{ssec:lexandlang}

The text used for training the acoustic model (Section \ref{ssec:AM})
was taken verbatim from the MALACH interview transcriptions in the
original LDC distribution. The transcriptions indicated disfluencies
and various types of noises. In these experiments, the noises were
eliminated from the transcripts with the assumption that during
training they would be
incorporated into the silence model automatically (which seemed to be
the case), but partial words and filled pauses were left as lexical
entries in the training text. 

One of the most challenging aspects of MALACH is the large number of
named entities, particularly of a foreign (non-US) nature. While
common words exist in any number of easily available pronunciation
lexicons (e.g., CMUDICT \cite{cmudict}), names, foreign words, and
partial words are not present. 
To create the MALACH
lexicon, a grapheme-to-phoneme system \cite{chen2003conditional}
followed by manual correction 
was utilized for those words not found in standard lexicons.

The language model was a 4-gram model created from the acoustic model training text
using modified Kneser-Ney smoothing
\cite{chen99}. Disfluencies were stripped out (informal experiments
suggested that blindly treating them as lexical entries hurt more than helped for recognition). Although earlier MALACH work \cite{ramabhadran2003towards} had reported gains
from  interpolation with other text sources, no such process was
performed in this work. The
perplexity of the minitest was 92.

\subsection{Acoustic Modeling}
\label{ssec:AM}

A set of acoustic models were built ranging from a basic
context-dependent hidden Markov model trained using a maximum
likelihood criterion up to a bi-directional LSTM model
employing multistream features and trained with a sMBR \cite{povey07} (state-level
minimum Bayes risk)
criterion. All training was done
utilizing  the IBM Attila toolkit \cite{soltau10} version 2.7 except
for the LSTM model. The LSTM model was trained using PyTorch \cite{pytorch2018}.
Neither data augmentation nor non-MALACH data was employed.

The Attila training recipes for context-dependent hidden Markov models, vocal
tract length normalization, feature space adaptation, and feature
space and model space MMI training are all described in
\cite{soltau10} and will not be reproduced here. The only important
thing to note is that the final decision tree consisted of 5000
context dependent states; no attempt was made to optimize this number
for best recognition performance.

The training for DNN and CNN hybrid models was also performed from a
native implementation in the IBM Attila toolkit. The inputs to the DNN
were nine-frame 40-dimensional PLP features after VTLN, LDA, and feature-space
normalization (FSA). The inputs to the CNN were eleven-frame 40-dimension logmel
features with deltas and delta-deltas after VTLN is
applied. Neural network configurations for both the DNN and CNN are
described in \cite{sainath2015deep}.
For the cross entropy (XE) training criterion, layerwise pre-training was followed by fine-tuning for
15 epochs on the entire network. The networks were
optimized using simple stochastic
gradient descent (SGD) with no particular bells or whistles. 
For the sequence training criterion, the Hessian-free (HF) optimization process
described in \cite{bedk12} was used starting from the fully trained
cross-entropy network.

The LSTM had the following configuration. The inputs were logmel
features (as above) augmented by delta and delta-delta features.
The network consisted of four bidirectional layers of 512 units
each, followed by a 256-unit linear projection layer into the 5000
context-dependent unit output layer. Training for the cross-entropy
criterion was done using Nesterov-based momentum with gradient clipping
and a droput factor of .25. The input was divided into minibatches of
256 21-element non-overlapping sequences each and trained on a single
GPU.

\subsection{Basic Speech Recognition Performance}
\label{ssec:performance}

All recognition results were obtained using the IBM Attila toolkit
\cite{soltau10}.
Table~\ref{tab:results1} displays Word Error Rate
(WER) results on the mini-devset.
All scoring
was performed using the NIST Scoring package {SCTK}-2.4.10 \cite{nistscore}.  
A global mapping (GLM) file (see {SCTK} documentation) was used to normalize spelling variants.
Disfluencies were marked as optional for scoring
(no penalty if deleted). 

\begin{table}[htp] 
\begin{center}
\begin{tabular}{|l|c|c|} \hline 
System              & MALACH          & 50-Hour \\
                    & minitest        & Broadcast News \\ \hline 
Context-Dependent   & 40.8            &  26.5                   \\ 
VTLN+FSA+MLLR       & 33.4            &  20.8                   \\
fMMI+BMMI+MLLR      & 29.8            &  15.5                   \\ 
DNN+XE              & 29.2            &  17.2                   \\ 
CNN+XE              & 28.7            &  15.9                   \\ 
DNN+HF              & 27.2            &  14.8                   \\ 
CNN+HF              & 26.8            &  13.7                   \\ 
LSTM                & 25.9            &  13.5                   \\ \hline            
\hspace{1mm}+splicing           & 25.4            &                         \\ 
\hspace{2mm}+sMBR               & 23.9            &                         \\ 
\hspace{3mm}+LSTM-LM            & 21.7            &                         \\ \hline            
\end{tabular} 
\end{center} 
\caption{Word error rates as a function of Acoustic Model for MALACH
  data and Broadcast News data.}
\vskip -0.4cm
\label{tab:results1}
\end{table}

Performance using simpler acoustic models is roughly similar to what was obtained
during the time of the original MALACH project when these sorts of
models were considered state-of-the-art. Speaker adaptation helps as
well as discriminative training in the form of fMMI and BMMI. Deep
learning-based models produce further performance improvements.
CNNs perform better than DNNs, and the sequence training
criterion produces better results than cross-entropy. The LSTM model
produces the overall best results (even without sequence training).

For comparison, results on the DEV-04f component using a 50-hour
training subset of Broadcast News (BN) data described in \cite{bedk09}
are included. The trends on BN are similar to MALACH but the performance
is appreciably better. This is no surprise insofar as the speakers are
largely professional announcers who are native speakers of American
English, but illustrates the challenge that speech
recognition still faces when presented with disfluent, emotional
speech from non-native English speakers.  

\section{Additional Recognition Results}
\label{sec:additional}

In \cite{saon2017english} one of the simpler but more successful
techniques to produce acoustic
model improvements was the application of feature fusion -
specifically, combining logmel features with feature-space adapted
features. To obtain better complementarity, a 64-dimensional filter bank
was created and used to extract logmel, delta, and delta-delta parameters.
Both sets of features were spliced into a
232-dimensional input vector. The system was then trained identically
to that of the LSTM system and obtained a WER of 25.4\% (vs 25.9\%) (Table~\ref{tab:results1}).

The models were then trained with the state-level sMBR
criterion using synchronous stochastic gradient with
momentum.
The numerator statistics for the sMBR training came from a
precomputed forced alignment of the training data, while the
denominator statistics came from lattices that are generated on
demand. To speed up training, parallel workers were used to compute the
gradients. A large number of utterances per batch were used to ensure
that reliable gradients are obtained, and because the gradients for
different batches are computed from a differing number of frames,
gradients are normalized by the number of frames prior to performing
parameter updates. The trainer is implemented using the PyTorch
\texttt{distributed} module. For the training runs in this paper, 
12 workers are used, 480-utterance batches, a learning rate of 1.0, and a
momentum of 0.9.  Only one epoch of training is performed because
additional epochs do not improve test performance of the acoustic
model. The resultant final word error rate was 23.9\% (Table~\ref{tab:results1}), a significant gain
over the 25.4\% reported for the LSTM on spliced parameters alone. 

Lastly, an LSTM based NN language model (NNLM) was trained.
Similar to the count models, the NNLM was trained on the acoustic transcription only (1.3M running words).
However, 10\% of the sentences were selected for cross-validation (CV) to control the learning rate schedule.
The NNLM has a word embedding layer with a size of 256, and three unidirectional LSTM layers, each with 512 nodes.
Before the softmax-based estimation of the 24k-dimensional posterior vector, the feature space was reduced to 128 by a linear bottleneck layer.
The model has 15M parameters.
In the field of small-scale language modeling, it is a well known
phenomenon that the best performing model has an order of magnitude more parameters than the number of available observations \cite{merity2018regularizing}.
In order to avoid over-fitting and co-adaptation of nodes, various dropout techniques were used \cite{JMLR:v15:srivastava14a,pmlr-v28-wan13}.
In each LSTM layer, DropConnect with a 30\% ratio was applied on the
hidden-to-hidden transformation matrix. In addition, 30\% of the outputs were also dropped out.
These two dropout parameters were set to 20\% in the embedding layer.
The initial learning rate was set to 0.01, and Nesterov momentum of 0.9 was also used.
After 30 epochs of training, the learning rate was annealed by a factor of $1/\sqrt{2}$ over 10 steps.
The final model has a perplexity of 70.8 on the CV set.
The lattices from the best LSTM acoustic model were generated using
the 4-gram LM and
rescored with this LSTM language model. The final error rate was
21.7\% (Table~\ref{tab:results1}), a significant drop in error rate
relative to the sMBR number of 23.9\%. 

\section{Discussion}
\label{sec:discussion}

As can be seen from the above results, although recent technology
advances have
made significant inroads in  performance,  MALACH remains a challenging
speech recognition task. To put things in perspective, 
the error rate on the popular Librispeech read speech corpus  when
trained using the 100-hour clean subset of the training data using a
DNN with p-norm and a heavily pruned language model is 9.19\% \cite{librispeech} (compared to the DNN-HF
on MALACH of 27.2\%); Broadcast News trained comparably (above) is 
at 13.5\% (vs. 25.7\% for MALACH), and one of the most difficult
public corpora containing relatively clean acoustic data,  the
close-talking microphone component of the spontaneous
multi-person AMI corpus \cite{carletta2006announcing}  is at 19.2\% \cite{amiresults}.

Much additional work needs to be done on this data to
establish a true state-of-the-art baseline for this task. All of the
above results represents a ``pure play'' on MALACH - no additional
data is being utilized. 
Early work
on MALACH \cite{ramabhadran2003towards} suggests that interpolation of MALACH text data with
other sources, such as Broadcast News and SWITCHBOARD, improves
performance.  It is also reasonable to believe that acoustic adaptation from a
much larger well-trained system might also produce better results than
just starting from scratch from the limited amount of MALACH training data.
Careful inspection of the test results revealed that a number of
issues still remained with respect to the
accuracy of the manual segmentation. Multiple additional verification passes are needed to really obtain a ``gold
standard'' reference script even for the minitest. Last, both the
minitest and the full test set contain
many sentence fragments making it more difficult for a long-span
language model such as an LSTM to really ``kick in'' to improve peformance; a
complete resegmentation (and expansion) of the test data is really
needed. 

\section{Summary}

The MALACH corpus is re-introduced as an important corpus because
of its societal interest and to challenge
the speech recognition community in areas such as modeling of accents,
disfluencies, and emotional speech. 
A range of systems were built spanning 
traditional HMMs all the way to hybrid LSTM-based acoustic and language models. 
The best system (trained purely on 176 hours of manually transcribed
speech and associated transcripts) presents a 21.7\% WER, compared
to the best results published during the original MALACH program of
32.1\% using 600 hours of transcribed and untranscribed data and a
language model interpolated with SWITCHBOARD and Broadcast News data. 
This demonstrates that while enormous strides in speech recognition have
been made, today's systems still have some distance to go before being able
to accurately transcribe difficult
data such as MALACH. To enable the community to continue research on
this important corpus, the training and test set definition, 
a reference lexicon, the GLM file, and other data useful for building and testing
speech recognition systems is now available from LDC \cite{LDC2019}.

\section{Acknowledgements}

We are grateful to our former colleague, Bhuvana Ramabhadran,
who did most of the original
speech recognition research on MALACH during the course of the MALACH 
project, including all the data preparation.
This work would not be possible without the support of Sam Gustman
from the Visual History Foundation. Thanks also goes to Doug Oard of
UMD for pushing hard for the public release of the data. This work
arises out of a five-year project originally funded by NSF under
the Information Technology Research (ITR) program, NSF IIS Award No. 0122466.

\bibliographystyle{IEEEbib}
\bibliography{Malach}

\end{document}